\documentclass[11pt]{article}

\usepackage{acl}

\usepackage{times}
\usepackage{latexsym}
\usepackage[T1]{fontenc}
\usepackage[utf8]{inputenc}
\usepackage{microtype}
\usepackage{inconsolata}
\usepackage{graphicx}
\usepackage{booktabs}
\usepackage{array}
\usepackage{multirow}
\usepackage{tabularx}
\usepackage{enumitem}
\usepackage{xcolor}
\usepackage{tcolorbox}
\tcbuselibrary{skins,breakable}
\usepackage{amsmath,amssymb}
\usepackage{url}
\usepackage{adjustbox}
\usepackage{placeins}
\usepackage{dblfloatfix}
\usepackage{float}

\graphicspath{{figures/}{image/}}

\newcommand{\code}[1]{\texttt{#1}}

\newcommand{\best}[1]{\textbf{#1}}

\newtcolorbox{errorblock}{
  colback=gray!4,
  colframe=gray!35,
  boxrule=0.4pt,
  arc=2pt,
  left=4pt,
  right=4pt,
  top=4pt,
  bottom=4pt,
  before skip=0.45em,
  after skip=0.45em
}

\newtcolorbox{promptbox}[1]{
  enhanced,
  breakable,
  colback=gray!2,
  colframe=gray!38,
  boxrule=0.45pt,
  arc=2pt,
  left=5pt,
  right=5pt,
  top=6pt,
  bottom=6pt,
  before skip=0.55em,
  after skip=0.65em,
  fontupper=\normalsize,
  fonttitle=\bfseries\normalsize,
  coltitle=black,
  title={#1},
  attach boxed title to top left={xshift=6pt,yshift=-2mm},
  boxed title style={colback=white,colframe=gray!38,boxrule=0.45pt,arc=2pt,left=4pt,right=4pt,top=1pt,bottom=1pt}
}

\newcommand{\promptrole}[1]{%
  \par\smallskip\noindent{\setlength{\fboxsep}{1.4pt}\colorbox{gray!16}{\small\textbf{#1}}}\quad
}

\newcommand{\promptsep}{\par\vspace{0.25em}\noindent\rule{\linewidth}{0.25pt}\vspace{-0.35em}}

\newtcolorbox{errorcard}[2][]{
  enhanced,
  breakable,
  colback=gray!2,
  colframe=gray!35,
  boxrule=0.45pt,
  arc=2pt,
  left=5pt,
  right=5pt,
  top=5pt,
  bottom=5pt,
  before skip=0.5em,
  after skip=0.55em,
  fontupper=\small,
  fonttitle=\bfseries\small,
  coltitle=black,
  title={#2},
  attach boxed title to top left={xshift=6pt,yshift=-2mm},
  boxed title style={colback=white,colframe=gray!35,boxrule=0.45pt,arc=2pt,left=4pt,right=4pt,top=1pt,bottom=1pt},
  #1
}

\newcommand{\errline}[2]{\par\noindent\textbf{#1.}\enspace #2}
\newcommand{\validex}[1]{\par\smallskip\noindent{\setlength{\fboxsep}{1.4pt}\colorbox{gray!12}{\scriptsize\textbf{Valid}}}\enspace #1}
\newcommand{\badex}[1]{\par\smallskip\noindent{\setlength{\fboxsep}{1.4pt}\colorbox{gray!18}{\scriptsize\textbf{Corrupted}}}\enspace #1}

\newtcolorbox{examplecard}[2][]{
  enhanced,
  breakable,
  colback=gray!2,
  colframe=gray!32,
  boxrule=0.4pt,
  arc=2pt,
  left=5pt,
  right=5pt,
  top=5pt,
  bottom=5pt,
  before skip=0.5em,
  after skip=0.6em,
  fontupper=\small,
  fonttitle=\bfseries\small,
  coltitle=black,
  title={#2},
  colbacktitle=gray!12,
  colframe=gray!32,
  #1
}

\newcommand{\stepfield}[2]{\par\noindent\makebox[5.8em][l]{{\scriptsize\bfseries\textsc{#1}}}#2}
\newcommand{\firsterror}[1]{\par\smallskip\noindent{\setlength{\fboxsep}{1.5pt}\colorbox{gray!20}{\scriptsize\bfseries First error}}\quad #1}

\title{Verifiable Counterfactual Supervision for Process Reward Models}

\author{Yinghui Chi \and Yuanhong Wang\thanks{Corresponding author.} \\
Jilin University}

\begin{document}
\maketitle

\begin{abstract}
Process reward models (PRMs) require supervision that identifies not only whether a reasoning trajectory is correct, but also where the reasoning process first becomes unsupported by its prefix. We frame this requirement as verifiable counterfactual process supervision with paired correct and erroneous trajectories in which the first invalid transition is known, the error mechanism is controlled, and the downstream continuation remains coherent under the corrupted state. Starting from a verified symbolic reasoning chain, our method injects a template-aware error at a selected intermediate step, recomputes all subsequent steps under the corrupted state, and verifies that the injected step is not derivable from its original prefix. The resulting trajectories provide prefix-valid first-error annotations and are translated into aligned natural-language processes for PRM training and evaluation. Experiments show that the synthesized data improve Best-of-8 reranking on logical reasoning benchmarks and show preliminary transfer to mathematical process evaluation. 
\end{abstract}

\section{Introduction}

Large language models (LLMs) have shown strong performance on multi-step reasoning tasks, especially when they are prompted to produce intermediate reasoning traces \citep{wei2022chain}. However, final-answer accuracy alone does not reveal whether the intermediate reasoning process is reliable. Process supervision addresses this limitation by providing step-level feedback, and has been shown to improve reasoning quality in complex settings \citep{uesato2022process,lightman2024verify}. In this setting, process reward models (PRMs) and step-level verifiers are increasingly used for generation, reranking, and search \citep{cobbe2021training,zhang2024genrm,setlur2024rewarding}.

The central challenge in process supervision is not simply obtaining more reasoning traces, but obtaining \emph{negative traces} with \emph{reliable} causal structure.
A negative reasoning trace is useful for PRM training only if it answers three questions: 
\emph{where} did the reasoning first go wrong, 
\emph{why} is that transition unsupported by the prefix, and 
\emph{how} should the subsequent trajectory evolve after the corrupted state is introduced?
Existing sources of process supervision only partially provide this structure.
Manual annotation can identify erroneous steps, but it is expensive and hard to scale~\citep{lightman2024verify}.
Rollout- and search-based annotation~\citep{wang2024mathshepherd,luo2024improve} reduce the cost, but they usually estimate step quality indirectly from final outcomes and offer limited control over the first-error position, error type, and downstream error propagation.

Identifying the first unsupported step in a reasoning trajectory requires knowing whether a step follows from its prefix. 
Symbolic systems provide exactly the machinery needed to test this derivability relation.
Recent work combined LLMs with symbolic provers to construct logically grounded reasoning data \citep{xu-etal-2024-faithful,leang-etal-2025-theorem,qi2025provergen}, which can be used for training and evaluating reasoning generation.
We utilize this line of work to instead synthesize negative process data as \emph{first-error counterfactuals}: trajectories that share a verified valid prefix with a correct chain, deviate at a controlled and verifiably unsupported step, and then continue under the corrupted state.

Our contributions are:
\begin{itemize}[leftmargin=*]
    \item We formulate verifiable counterfactual supervision for PRMs, where each negative trajectory has a known prefix-unsupported transition.
    \item We propose a symbolic synthesis pipeline that combines template-aware error injection, prefix non-derivability checking, and downstream recomputation under the corrupted state.
    \item We show that the resulting data improves Best-of-8 reranking on logical reasoning benchmarks and provides a diagnostic testbed for first-error localization.
\end{itemize}


\section{Related Work}

Process supervision trains models to evaluate intermediate reasoning steps rather than only final answers \citep{uesato2022process,lightman2024verify}. PRMs have been used for reranking, search, and reasoning-time selection \citep{cobbe2021training,zhang2024genrm,setlur2024rewarding}. However, recent benchmarks show that fine-grained process evaluation, especially first-error localization, remains difficult \citep{zheng2024processbench,song2025prmbench}.

Manual annotation provides reliable step labels but is expensive \citep{lightman2024verify}. Rollout- and search-based methods reduce annotation cost \citep{wang2024mathshepherd,luo2024improve}, but usually infer step quality indirectly from final outcomes and offer limited control over error type, error position, and downstream consistency. Direct synthesis is more scalable \citep{zelikman2022star,zeng2025versaprm}, but existing methods generally do not enforce verifiable first-error counterfactual structure.


Recent work combines LLMs with symbolic provers to construct logically grounded reasoning data \citep{xu-etal-2024-faithful,leang-etal-2025-theorem,qi2025provergen}. These methods primarily focus on verified correct chains. In contrast, we use symbolic verification to synthesize controlled erroneous chains for PRM supervision.


\begin{figure*}[!t]
  \centering
  \includegraphics[width=0.8\textwidth]{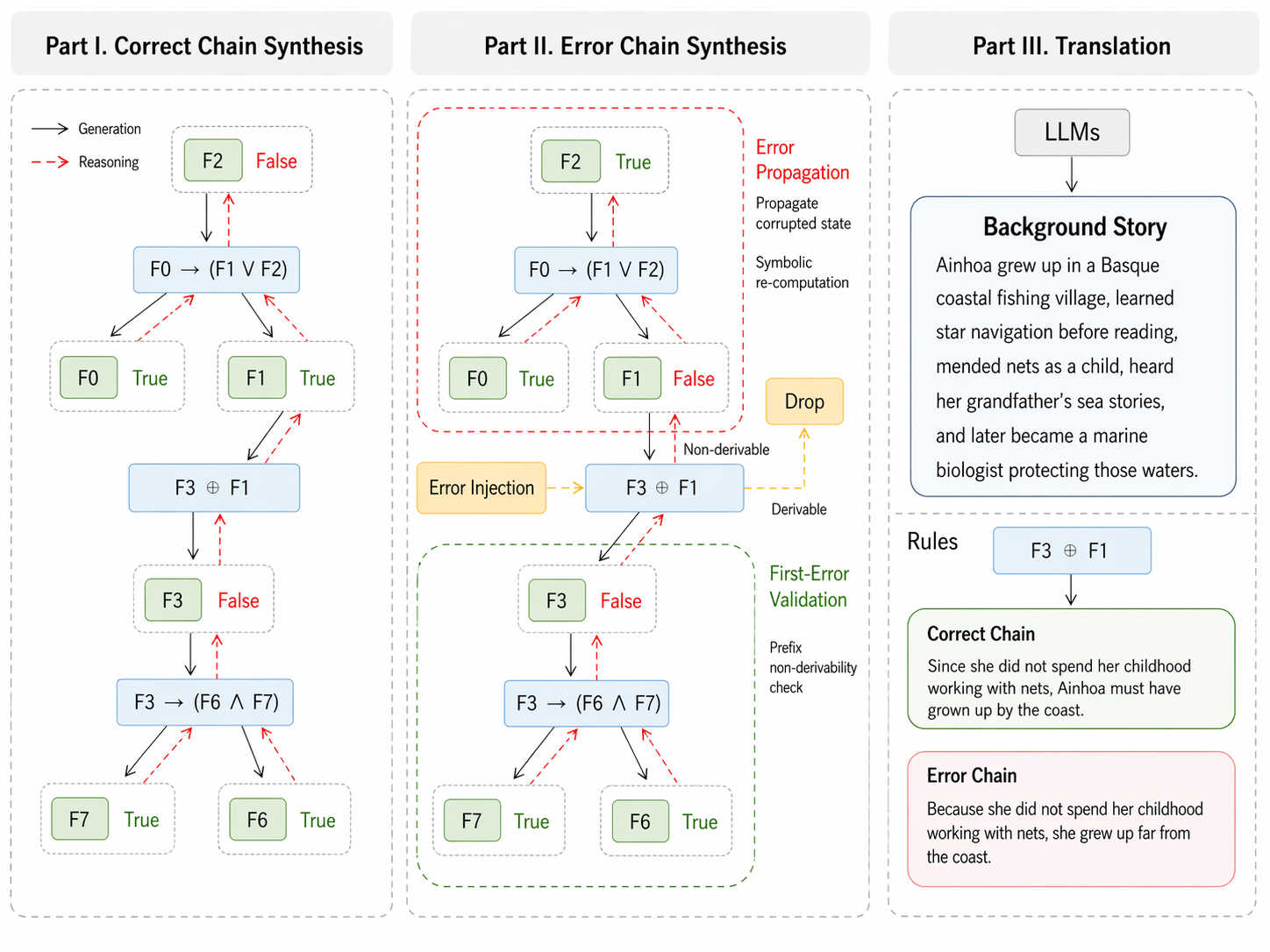}
  \caption{Overview of the proposed verifiable counterfactual process supervision framework for PRMs. 
  }
  \label{fig:framework}
\end{figure*}

\section{Method}

We first formalize the notion of first-error counterfactual supervision for PRMs, and then describe our synthesis framework as illustrated in Figure~\ref{fig:framework}.

\subsection{Problem Formulation}

Let $C=(s_1,\ldots,s_T)$ be a reasoning trajectory, where each step $s_t$ consists of a set of supporting facts, an applied rule, and a conclusion. Let $\mathcal{P}_t$ denote the prefix state induced by the initial facts and previous steps $s_{<t}$. We say that $s_t$ is \emph{prefix-valid} if its conclusion is derivable from $\mathcal{P}_t$ under the rule set.

A first-error counterfactual instance is a tuple $(G, C^+, C^-, k, e)$, where $G$ is the target goal, $C^+$ is a verified correct chain, $C^-$ is a corrupted chain, $k$ is the first-error position, and $e$ is the error type. The corrupted chain should satisfy $s^-_t = s^+_t \quad \text{for all } t < k$, $s^-_k$ is prefix-invalid with respect to $\mathcal{P}_k$ and error type $e$, and downstream steps $s^-_{>k}$ are recomputed under the corrupted state induced by $s^-_k$.
Thus, the corrupted trajectory is invalid with respect to the original problem state, but remains coherent as a counterfactual continuation after the injected error.



\subsection{Correct Chain Synthesis}

We first generate a correct symbolic reasoning chain from a rule template pool. Starting from the target conclusion, we recursively expand intermediate goals and use a symbolic prover to verify whether each step is derivable from the current premises. Only truth-consistent structures are retained, yielding a verifiable forward chain from base facts to the target.

\subsection{Error Injection and Symbolic Re-computation}

Given a verified correct chain $C^+ = (s_1, \dots, s_T)$, we construct an erroneous counterpart by first selecting an intermediate step $k$ as the candidate first-error position. The perturbation type $e$ is constrained by the rule template instantiated at step $k$, ensuring that the injected error is structurally compatible with the local inference pattern rather than sampled from a global error distribution. Depending on the template, the perturbation may alter the inferred conclusion, misuse the rule direction, or corrupt the local factual state.

The corrupted conclusion produced at step $k$ is then treated as part of the subsequent symbolic state. We recompute all downstream steps under this updated state with the symbolic prover. 
The resulting erroneous chain $C^-$ preserves the correct prefix before step $k$, replaces $s_k$ with a corrupted step, and recomputes all subsequent steps from the corrupted state. 
It is essentially a counterfactual trajectory that shares the same initial conditions and rule applications as the original chain, but with a controlled error injected at step $k$ and propagated through the rest of the trajectory.

To verify that the injected step is indeed the first invalid step, we apply error-type-specific checks. For non-structural errors, we perform a prefix non-derivability check by testing whether the corrupted conclusion $\tilde{s}_k$ is derivable from the original prefix $(s_1, \dots, s_{k-1})$, and the sample is rejected if it is still derivable. 
For structural errors, we instead verify that the corrupted step violates the intended rule direction, dependency order, or acyclic trajectory constraints, since the local truth values may still be compatible with the instantiated rule. Detailed error type distributions and definitions are provided in Appendices~\ref{app:error-distribution} and~\ref{app:error-definitions}.

\subsection{Natural-Language Translation}

Finally, we convert the symbolic trajectories into natural-language reasoning processes. To preserve comparability between the correct and erroneous chains, both translations are conditioned on the same background context, entity realizations, and predicate mappings. This shared semantic grounding ensures that the two processes differ primarily in their reasoning behavior rather than in unrelated surface realizations.


\section{Experiments}

\subsection{Experimental Setup}

We evaluate our approach through three settings: Best-of-$K$ reranking for logical reasoning, transfer evaluation on mathematical reasoning, and step-level evaluation on our synthesized $20$K process supervision data.

We use Llama-3.1-8B~\citep{grattafiori2024llama} and Qwen-2.5-7B~\citep{yang2024qwen25}. 
For Best-of-8 reranking, the candidate solutions are generated by the corresponding base model and then selected by a scorer. 
Thus, improvements reflect better trajectory selection rather than stronger candidate generation.

For comparison, we include domain-adapted baselines Llama-3.1-8B-D and Llama-3.1-8B-M trained on DeepSeek-style and Mistral-style process data\citep{guo2025deepseekr1,jiang2023mistral}, as well as Qwen-2.5-7B-math and Qwen-2.5-7B-math-800K, which are publicly released math-adapted variants \citep{yang2024qwen25math}. 
More details are provided in the Appendix~\ref{app:dataset-eval}.

\subsection{Main Results on Logical Reasoning}

\begin{table*}[!t]
\centering
\caption{Best-of-8 reranking results on logical reasoning benchmarks. Candidate solutions are generated by the Llama-3.1-8B and Qwen-2.5-7B base models, respectively. Best results are bolded.}
\label{tab:main}
\small
\resizebox{\textwidth}{!}{%
\begin{tabular}{lcccccccc}
\toprule
\textbf{PRM} & \textbf{folio} & \textbf{logicnli} & \textbf{pro\_nomath} & \textbf{anli} & \textbf{temporal} & \textbf{tracking} & \textbf{hans} & \textbf{avg} \\
\midrule
llama-3.1-8B & 0.575 & 0.472 & 0.498 & 0.729 & 0.412 & 0.391 & 0.620 & 0.528 \\
\midrule
llama-3.1-8B-D & 0.519 & 0.489 & 0.528 & 0.748 & \best{0.611} & \best{0.488} & 0.644 & 0.575 \\
llama-3.1-8B-M & 0.516 & \best{0.490} & 0.491 & 0.752 & 0.598 & 0.384 & 0.670 & 0.557 \\
\midrule
\best{llama-prm (ours)} & \best{0.625} & 0.473 & \best{0.584} & \best{0.772} & 0.569 & 0.396 & \best{0.721} & \best{0.591} \\
\midrule\midrule
qwen-2.5-7B & 0.543 & 0.456 & 0.409 & 0.655 & 0.659 & 0.543 & 0.710 & 0.567 \\
\midrule
qwen-2.5-7B-math & 0.512 & 0.494 & 0.445 & \best{0.701} & 0.681 & \best{0.667} & 0.771 & 0.610 \\
qwen-2.5-7B-math-800K & 0.524 & 0.514 & 0.442 & 0.655 & 0.679 & 0.635 & 0.763 & 0.601 \\
\midrule
\best{qwen-prm (ours)} & \best{0.584} & \best{0.527} & \best{0.458} & 0.691 & \best{0.712} & 0.562 & \best{0.780} & \best{0.615} \\
\bottomrule
\end{tabular}}
\end{table*}

Tables~\ref{tab:main} report Best-of-8 reranking results on logical reasoning benchmarks. 
Across both model families, our PRM improves the average score over the corresponding base selector. 
For Llama-3.1-8B, the average score increases from $0.528$ to $0.591$, and for Qwen-2.5-7B, it increases from $0.567$ to $0.615$. 
Compared with the strongest domain-adapted baseline in each family, our PRM still obtains the best average performance.

The improvements are greatest on datasets such as folio, pro-nomath, anli, and hans for Llama, and folio, logicnli, pro-nomath, temporal, and hans for Qwen. 
At the same time, domain-adapted baselines remain stronger on some tasks, such as tracking. 
These results suggest that first-error counterfactual supervision learns a selection signal that might be complementary to domain-adapted process data. 
We additionally compare against Oracle@8 in Appendix~\ref{app:best-of-8}, where the gap indicates that process-level supervision still has room for improvement in selecting the best candidate.

\subsection{Transfer to Mathematical Reasoning}

We further test whether the process-level signal learned from symbolic trajectories transfers to a different reasoning domain. 
Table~\ref{tab:math-transfer} reports F1 scores on the mathematical subsets of ProcessBench~\citep{zheng2024processbench}. 
Although the PRM is trained on synthesized symbolic process data rather than mathematical derivations, increasing the amount of synthesized data consistently improves performance.
The average F1 rises from $4.1$ for the base model to $10.0$ with $5$K instances and $19.2$ with $20$K instances.
The absolute scores remain low, but the monotonic improvement from 5K to 20K suggests that prefix-validity supervision may provide a weak but nontrivial signal beyond the synthetic symbolic domain.

\begin{table}[!t]
\centering
\caption{Transfer evaluation (F1 scores) on mathematical reasoning tasks in ProcessBench. 
}
\label{tab:math-transfer}
\small
\setlength{\tabcolsep}{4pt}
\resizebox{\columnwidth}{!}{%
\begin{tabular}{lccccc}
\toprule
\textbf{Model} & \textbf{gsm8k} & \textbf{math} & \textbf{olymp.} & \textbf{omni.} & \textbf{avg} \\
\midrule
llama-3.1-8B & 3.8 & 3.0 & 5.3 & 4.1 & 4.1 \\
llama-3.1-8B-5k & 18.0 & 10.3 & 4.1 & 7.9 & 10.0 \\
\midrule
\best{llama-3.1-8B-20k} & \best{21.2} & \best{21.3} & \best{18.6} & \best{16.0} & \best{19.2} \\
\bottomrule
\end{tabular}
}
\end{table}

\subsection{Step-level Evaluation on Our Data}

\begin{table}[!t]
\centering
\caption{Step-level evaluation on our process supervision data.
}
\label{tab:step-eval}
\small
\begin{tabular}{lcc}
\toprule
\textbf{Model} & \textbf{first\_error} & \textbf{all\_step} \\
\midrule
gpt-4o & 0.160 & 0.479 \\
gpt-5 & 0.575 & 0.866 \\
kimi-k2.5 & 0.103 & 0.632 \\
DeepSeek-V3.2 & 0.333 & 0.608 \\
gemini-3.1 & 0.825 & 0.941 \\
\bottomrule
\end{tabular}
\end{table}

Table~\ref{tab:step-eval} evaluates models on the fine-grained annotations provided by our synthesized data. 
Each model is tested on $100$ synthesized trajectories with $7$--$10$ reasoning steps. 
We report \texttt{all\_step} measuring whether the model correctly classifies each step as valid or invalid, and \texttt{first\_error} measuring whether the model correctly identifies the first unsupported step in the trajectory.

The results show a large gap between these two abilities. 
Even the strongest evaluated model, Gemini-3.1, has a lower first-error accuracy ($0.825$) than all-step accuracy ($0.941$). 
This confirms that first-error localization is not merely ordinary step classification.
A model recognizing individual steps as plausible or implausible may still fail to identify the earliest unsupported transition.

\section{Conclusion}

We introduced verifiable first-error counterfactual supervision for process reward models. 
Rather than treating negative process data as independently labeled erroneous steps, our framework constructs paired correct and corrupted trajectories in which the first unsupported transition is controlled, verified, and propagated through a coherent counterfactual continuation. 
This design turns symbolic verification into a supervision mechanism for PRMs. 
Experiments on logical reasoning benchmarks show that the resulting data improves Best-of-8 reranking, and our diagnostic evaluation suggests that first-error localization remains substantially more challenging than ordinary step-level validity classification. 
These findings indicate that reliable negative process supervision is a useful direction for building more fine-grained reasoning evaluators. 

\clearpage

\section*{Limitations}

While our framework improves controllability and verifiability, it has several limitations. First, the diversity of synthesized errors is still limited by the predefined templates and may not fully cover the range of errors in open-ended LLM reasoning. In addition, the current framework is built on symbolic reasoning structures, which makes it most naturally applicable to domains with relatively explicit logical form. Finally, while the current results are encouraging, evaluating the framework on broader tasks and model families would further strengthen its generality.

\section*{Acknowledgments}

Yuanhong Wang is supported by the National Natural Science Foundation of China under Grant No. 62506141. We thank the reviewers and colleagues for their helpful feedback.

\FloatBarrier
\bibliography{refs}

@article{cobbe2021training,
  title = {Training Verifiers to Solve Math Word Problems},
  author = {Cobbe, Karl and Kosaraju, Vineet and Bavarian, Mohammad and Chen, Mark and Jun, Heewoo and Kaiser, Lukasz and Plappert, Matthias and Tworek, Jerry and Hilton, Jacob and Nakano, Reiichiro and Hesse, Christopher and Schulman, John},
  journal = {arXiv preprint arXiv:2110.14168},
  year = {2021},
  doi = {10.48550/arXiv.2110.14168},
  url = {https://arxiv.org/abs/2110.14168}
}

@article{uesato2022process,
  title = {Solving Math Word Problems with Process- and Outcome-Based Feedback},
  author = {Uesato, Jonathan and Kushman, Nate and Kumar, Ramana and Song, Francis and Siegel, Noah and Wang, Lisa and Creswell, Antonia and Irving, Geoffrey and Higgins, Irina},
  journal = {arXiv preprint arXiv:2211.14275},
  year = {2022},
  doi = {10.48550/arXiv.2211.14275},
  url = {https://arxiv.org/abs/2211.14275}
}

@inproceedings{lightman2024verify,
  title = {Let's Verify Step by Step},
  author = {Lightman, Hunter and Kosaraju, Vineet and Burda, Yura and Edwards, Harri and Baker, Bowen and Lee, Teddy and Leike, Jan and Schulman, John and Sutskever, Ilya and Cobbe, Karl},
  booktitle = {The Twelfth International Conference on Learning Representations},
  year = {2024},
  url = {https://openreview.net/forum?id=v8L0pN6EOi}
}

@inproceedings{wang2024mathshepherd,
  title = {Math-Shepherd: Verify and Reinforce {LLMs} Step-by-step without Human Annotations},
  author = {Wang, Peiyi and Li, Lei and Shao, Zhihong and Xu, Runxin and Dai, Damai and Li, Yifei and Chen, Deli and Wu, Yu and Sui, Zhifang},
  booktitle = {Proceedings of the 62nd Annual Meeting of the Association for Computational Linguistics (Volume 1: Long Papers)},
  pages = {9426--9439},
  address = {Bangkok, Thailand},
  publisher = {Association for Computational Linguistics},
  year = {2024},
  url = {https://aclanthology.org/2024.acl-long.510/},
  doi = {10.18653/v1/2024.acl-long.510}
}

@inproceedings{zheng2024processbench,
  title = {{P}rocess{B}ench: Identifying Process Errors in Mathematical Reasoning},
  author = {Zheng, Chujie and Zhang, Zhenru and Zhang, Beichen and Lin, Runji and Lu, Keming and Yu, Bowen and Liu, Dayiheng and Zhou, Jingren and Lin, Junyang},
  booktitle = {Proceedings of the 63rd Annual Meeting of the Association for Computational Linguistics (Volume 1: Long Papers)},
  pages = {1009--1024},
  address = {Vienna, Austria},
  publisher = {Association for Computational Linguistics},
  year = {2025},
  url = {https://aclanthology.org/2025.acl-long.50/},
  doi = {10.18653/v1/2025.acl-long.50}
}

@inproceedings{song2025prmbench,
  title = {{PRMB}ench: A Fine-grained and Challenging Benchmark for Process-Level Reward Models},
  author = {Song, Mingyang and Su, Zhaochen and Qu, Xiaoye and Zhou, Jiawei and Cheng, Yu},
  booktitle = {Proceedings of the 63rd Annual Meeting of the Association for Computational Linguistics (Volume 1: Long Papers)},
  pages = {25299--25346},
  address = {Vienna, Austria},
  publisher = {Association for Computational Linguistics},
  year = {2025},
  url = {https://aclanthology.org/2025.acl-long.1230/},
  doi = {10.18653/v1/2025.acl-long.1230}
}

@inproceedings{qi2025provergen,
  title = {Large Language Models Meet Symbolic Provers for Logical Reasoning Evaluation},
  author = {Qi, Chengwen and Ma, Ren and Li, Bowen and Du, He and Hui, Binyuan and Wu, Jinwang and Laili, Yuanjun and He, Conghui},
  booktitle = {The Thirteenth International Conference on Learning Representations},
  year = {2025},
  url = {https://openreview.net/forum?id=C25SgeXWjE}
}

@article{zeng2025versaprm,
  title = {{VersaPRM}: Multi-Domain Process Reward Model via Synthetic Reasoning Data},
  author = {Zeng, Thomas and Zhang, Shuibai and Wu, Shutong and Classen, Christian and Chae, Daewon and Ewer, Ethan and Lee, Minjae and Kim, Heeju and Kang, Wonjun and Kunde, Jackson and Fan, Ying and Kim, Jungtaek and Koo, Hyung Il and Ramchandran, Kannan and Papailiopoulos, Dimitris and Lee, Kangwook},
  journal = {arXiv preprint arXiv:2502.06737},
  year = {2025},
  url = {https://arxiv.org/abs/2502.06737},
  doi = {10.48550/arXiv.2502.06737}
}

@article{grattafiori2024llama,
  title = {The {Llama 3} Herd of Models},
  author = {Grattafiori, Aaron and Dubey, Abhimanyu and Jauhri, Abhinav and Pandey, Abhinav and Kadian, Abhishek and Al-Dahle, Ahmad and Letman, Aiesha and Mathur, Akhil and Schelten, Alan and Vaughan, Alex and Yang, Amy and Fan, Angela},
  journal = {arXiv preprint arXiv:2407.21783},
  year = {2024},
  url = {https://arxiv.org/abs/2407.21783}
}

@article{yang2024qwen25,
  title = {{Qwen2.5} Technical Report},
  author = {{Qwen Team}},
  journal = {arXiv preprint arXiv:2412.15115},
  year = {2024},
  url = {https://arxiv.org/abs/2412.15115}
}

@article{yang2024qwen25math,
  title = {{Qwen2.5-Math} Technical Report: Toward Mathematical Expert Model via Self-Improvement},
  author = {Yang, An and Zhang, Beichen and Hui, Binyuan and Gao, Bofei and Yu, Bowen and Li, Chengpeng and Liu, Dayiheng and Tu, Jianhong and Zhou, Jingren and Lin, Junyang and Lu, Keming and Xue, Mingfeng and Lin, Runji and Liu, Tianyu and Ren, Xingzhang and Zhang, Zhenru},
  journal = {arXiv preprint arXiv:2409.12122},
  year = {2024},
  url = {https://arxiv.org/abs/2409.12122}
}

@inproceedings{han2022folio,
  title = {{FOLIO}: Natural Language Reasoning with First-Order Logic},
  author = {Han, Simeng and Schoelkopf, Hailey and Zhao, Yilun and Qi, Zhenting and Riddell, Martin and Zhou, Wenfei and Coady, James and Peng, David and Qiao, Yujie and Benson, Luke and Sun, Lucy and Wardle-Solano, Alexander and Szab{\'o}, Hannah and Zubova, Ekaterina and Burtell, Matthew and Fan, Jonathan and Liu, Yixin and Wong, Brian},
  booktitle = {Proceedings of the 2024 Conference on Empirical Methods in Natural Language Processing},
  year = {2024},
  pages = {22017--22031},
  address = {Miami, Florida, USA},
  publisher = {Association for Computational Linguistics},
  url = {https://aclanthology.org/2024.emnlp-main.1229/},
  doi = {10.18653/v1/2024.emnlp-main.1229}
}

@inproceedings{tian2021logicnli,
  title = {Diagnosing the First-Order Logical Reasoning Ability Through {LogicNLI}},
  author = {Tian, Jidong and Li, Yitian and Chen, Wenqing and Xiao, Liqiang and He, Hao and Jin, Yaohui},
  booktitle = {Proceedings of the 2021 Conference on Empirical Methods in Natural Language Processing},
  year = {2021},
  pages = {3738--3747},
  address = {Online and Punta Cana, Dominican Republic},
  publisher = {Association for Computational Linguistics},
  url = {https://aclanthology.org/2021.emnlp-main.303/},
  doi = {10.18653/v1/2021.emnlp-main.303}
}

@inproceedings{nie2020anli,
  title = {Adversarial {NLI}: A New Benchmark for Natural Language Understanding},
  author = {Nie, Yixin and Williams, Adina and Dinan, Emily and Bansal, Mohit and Weston, Jason and Kiela, Douwe},
  booktitle = {Proceedings of the 58th Annual Meeting of the Association for Computational Linguistics},
  year = {2020},
  pages = {4885--4901},
  address = {Online},
  publisher = {Association for Computational Linguistics},
  url = {https://aclanthology.org/2020.acl-main.441/},
  doi = {10.18653/v1/2020.acl-main.441}
}

@inproceedings{mccoy2019hans,
  title = {Right for the Wrong Reasons: Diagnosing Syntactic Heuristics in Natural Language Inference},
  author = {McCoy, R. Thomas and Pavlick, Ellie and Linzen, Tal},
  booktitle = {Proceedings of the 57th Annual Meeting of the Association for Computational Linguistics},
  year = {2019},
  pages = {3428--3448},
  address = {Florence, Italy},
  publisher = {Association for Computational Linguistics},
  url = {https://aclanthology.org/P19-1334/},
  doi = {10.18653/v1/P19-1334}
}

@article{Suzgun2023BBH,
  title = {Challenging {BIG-Bench} Tasks and Whether Chain-of-Thought Can Solve Them},
  author = {Suzgun, Mirac and Scales, Nathan and Sch{\"a}rli, Nathanael and Gehrmann, Sebastian and Tay, Yi and Chung, Hyung Won and Chowdhery, Aakanksha and Le, Quoc V. and Chi, Ed H. and Zhou, Denny and Wei, Jason},
  journal = {arXiv preprint arXiv:2210.09261},
  year = {2022},
  url = {https://arxiv.org/abs/2210.09261}
}

@inproceedings{wang2024mmlupro,
  title = {{MMLU-Pro}: A More Robust and Challenging Multi-Task Language Understanding Benchmark},
  author = {Wang, Yubo and Ma, Xueguang and Zhang, Ge and Ni, Yuansheng and Chandra, Abhranil and Guo, Shiguang and Ren, Weiming and Arulraj, Aaran and He, Xuan and Jiang, Ziyan and Li, Tianle and Ku, Max and Wang, Kai and Zhuang, Alex and Fan, Rongqi and Yue, Xiang and Chen, Wenhu},
  booktitle = {Advances in Neural Information Processing Systems},
  year = {2024},
  url = {https://arxiv.org/abs/2406.01574}
}

@article{hendrycks2021math,
  title = {Measuring Mathematical Problem Solving With the {MATH} Dataset},
  author = {Hendrycks, Dan and Burns, Collin and Kadavath, Saurav and Arora, Akul and Basart, Steven and Tang, Eric and Song, Dawn and Steinhardt, Jacob},
  journal = {arXiv preprint arXiv:2103.03874},
  year = {2021},
  url = {https://arxiv.org/abs/2103.03874}
}

@article{he2024olympiadbench,
  title = {{OlympiadBench}: A Challenging Benchmark for Promoting {AGI} with Olympiad-Level Bilingual Multimodal Scientific Problems},
  author = {He, Chaoqun and Luo, Renjie and Bai, Yuzhuo and Hu, Shengding and Thai, Zhen Leng and Shen, Junhao and Hu, Jinyi and Han, Xu and Huang, Yujie and Zhang, Yuxiang and Liu, Jie and Qi, Lei and Liu, Zhiyuan and Sun, Maosong},
  journal = {arXiv preprint arXiv:2402.14008},
  year = {2024},
  url = {https://arxiv.org/abs/2402.14008}
}

@article{gao2024omnimath,
  title = {{Omni-MATH}: A Universal Olympiad Level Mathematic Benchmark for Large Language Models},
  author = {Gao, Bofei and Song, Feifan and Yang, Zhe and Cai, Zefan and Miao, Yibo and Dong, Qingxiu and Li, Lei and Ma, Chenghao and Chen, Liang and Xu, Runxin and Tang, Zhengyang and Wang, Benyou and Zan, Daoguang and Quan, Shanghaoran and Zhang, Ge and Sha, Lei and Zhang, Yichang and Ren, Xuancheng and Liu, Tianyu and Chang, Baobao},
  journal = {arXiv preprint arXiv:2410.07985},
  year = {2024},
  url = {https://arxiv.org/abs/2410.07985}
}

@inproceedings{wei2022chain,
  title = {Chain-of-Thought Prompting Elicits Reasoning in Large Language Models},
  author = {Wei, Jason and Wang, Xuezhi and Schuurmans, Dale and Bosma, Maarten and Ichter, Brian and Xia, Fei and Chi, Ed H. and Le, Quoc V. and Zhou, Denny},
  booktitle = {Advances in Neural Information Processing Systems},
  volume = {35},
  pages = {24824--24837},
  year = {2022},
  url = {https://proceedings.neurips.cc/paper_files/paper/2022/hash/9d5609613524ecf4f15af0f7b31abca4-Abstract-Conference.html}
}

@article{setlur2024rewarding,
  title = {Rewarding Progress: Scaling Automated Process Verifiers for {LLM} Reasoning},
  author = {Setlur, Amrith and Nagpal, Chirag and Fisch, Adam and Geng, Xinyang and Eisenstein, Jacob and Agarwal, Rishabh and Agarwal, Alekh and Berant, Jonathan and Kumar, Aviral},
  journal = {arXiv preprint arXiv:2410.08146},
  year = {2024},
  url = {https://arxiv.org/abs/2410.08146}
}

@article{luo2024improve,
  title = {Improve Mathematical Reasoning in Language Models by Automated Process Supervision},
  author = {Luo, Liangchen and Liu, Yinxiao and Liu, Rosanne and Phatale, Samrat and Guo, Meiqi and Lara, Harsh and Li, Yunxuan and Shu, Lei and Zhu, Yun and Meng, Lei and Sun, Jiao and Rastogi, Abhinav},
  journal = {arXiv preprint arXiv:2406.06592},
  year = {2024},
  url = {https://arxiv.org/abs/2406.06592}
}

@article{zhang2024genrm,
  title = {Generative Verifiers: Reward Modeling as Next-Token Prediction},
  author = {Zhang, Lunjun and Hosseini, Arian and Bansal, Hritik and Kazemi, Mehran and Kumar, Aviral and Agarwal, Rishabh},
  journal = {arXiv preprint arXiv:2408.15240},
  year = {2024},
  url = {https://arxiv.org/abs/2408.15240}
}

@inproceedings{zelikman2022star,
  title = {{STaR}: Bootstrapping Reasoning with Reasoning},
  author = {Zelikman, Eric and Wu, Yuhuai and Mu, Jesse and Goodman, Noah D.},
  booktitle = {Advances in Neural Information Processing Systems},
  volume = {35},
  pages = {15476--15488},
  year = {2022},
  url = {https://arxiv.org/abs/2203.14465}
}

@inproceedings{xu-etal-2024-faithful,
    title = "Faithful Logical Reasoning via Symbolic Chain-of-Thought",
    author = "Xu, Jundong and Fei, Hao and Pan, Liangming and Liu, Qian and Lee, Mong-Li and Hsu, Wynne",
    editor = "Ku, Lun-Wei and Martins, Andre and Srikumar, Vivek",
    booktitle = "Proceedings of the 62nd Annual Meeting of the Association for Computational Linguistics (Volume 1: Long Papers)",
    month = aug,
    year = "2024",
    address = "Bangkok, Thailand",
    publisher = "Association for Computational Linguistics",
    url = "https://aclanthology.org/2024.acl-long.720/",
    doi = "10.18653/v1/2024.acl-long.720",
    pages = "13326--13365"
}

@inproceedings{leang-etal-2025-theorem,
    title = "Theorem Prover as a Judge for Synthetic Data Generation",
    author = "Leang, Joshua Ong Jun and
      Hong, Giwon and
      Li, Wenda and
      Cohen, Shay B.",
    editor = "Che, Wanxiang and
      Nabende, Joyce and
      Shutova, Ekaterina and
      Pilehvar, Mohammad Taher",
    booktitle = "Proceedings of the 63rd Annual Meeting of the Association for Computational Linguistics (Volume 1: Long Papers)",
    month = jul,
    year = "2025",
    address = "Vienna, Austria",
    publisher = "Association for Computational Linguistics",
    url = "https://aclanthology.org/2025.acl-long.1448/",
    doi = "10.18653/v1/2025.acl-long.1448",
    pages = "29941--29977",
    ISBN = "979-8-89176-251-0"
}

@article{guo2025deepseekr1,
  title = {DeepSeek-R1: Incentivizing Reasoning Capability in LLMs via Reinforcement Learning},
  author = {Guo, Daya and Yang, Dejian and Zhang, Haowei and Song, Junxiao and Zhang, Ruoyu and Xu, Runxin and Zhu, Qihao and Ma, Shirong and Wang, Peiyi and Bi, Xiao and others},
  journal = {arXiv preprint arXiv:2501.12948},
  year = {2025},
  url = {https://arxiv.org/abs/2501.12948}
}

@article{jiang2023mistral,
  title = {Mistral 7B},
  author = {Jiang, Albert Q. and Sablayrolles, Alexandre and Mensch, Arthur and Bamford, Chris and Chaplot, Devendra Singh and de las Casas, Diego and Bressand, Florian and Lengyel, Gianna and Lample, Guillaume and Saulnier, Lucile and Lavaud, L{\'e}lio Renard and Lachaux, Marie-Anne and Stock, Pierre and Le Scao, Teven and Lavril, Thibaut and Wang, Thomas and Lacroix, Timoth{\'e}e and El Sayed, William},
  journal = {arXiv preprint arXiv:2310.06825},
  year = {2023},
  url = {https://arxiv.org/abs/2310.06825}
}

\clearpage

\appendix

\section{Dataset and Evaluation Details}
\label{app:dataset-eval}
This appendix provides additional implementation and evaluation details to support reproducibility.

For logical reasoning evaluation, we use FOLIO \citep{han2022folio}, LogicNLI \citep{tian2021logicnli}, ANLI \citep{nie2020anli}, and HANS \citep{mccoy2019hans}, together with three additional logic-oriented tasks: Pro-NoMath \citep{wang2024mmlupro}, Temporal, and Tracking \citep{Suzgun2023BBH}. These datasets cover different forms of reasoning, including first-order logical inference, natural language inference, heuristic-sensitive inference, temporal reasoning, and object/state tracking. For transfer evaluation, we use GSM8K \citep{cobbe2021training}, MATH \citep{hendrycks2021math}, OlympiadBench \citep{he2024olympiadbench}, and Omni-MATH \citep{gao2024omnimath}. The logical reasoning tasks are used for Best-of-8 reranking, while the mathematical reasoning tasks are used to test whether the learned process-level signal transfers beyond the symbolic setting used for data synthesis.

\paragraph{Dataset licenses and usage.}
All external datasets used in this work are public research benchmarks and are used only for offline evaluation or Best-of-8 reranking. We do not use their raw examples in our synthesized PRM training data and do not redistribute them. We cite the original papers and list the public repositories in Table~\ref{tab:dataset-details}. License or access-term information was checked from public repository metadata when available; for datasets without clearly specified license information, we restrict our use to research evaluation through the original access channel. No private user data or newly collected personal information is introduced by our preprocessing.

\begin{table}[H]
\centering
\small
\setlength{\tabcolsep}{3pt}
\begin{tabularx}{\columnwidth}{@{}l l X@{}}
\toprule
\textbf{Dataset} & \textbf{Use} & \textbf{Repository} \\
\midrule
FOLIO & Logic & \href{https://huggingface.co/datasets/tasksource/folio}{\nolinkurl{tasksource/folio}} \\
LogicNLI & Logic & \href{https://huggingface.co/datasets/tasksource/LogicNLI}{\nolinkurl{tasksource/LogicNLI}} \\
ANLI & Logic & \href{https://huggingface.co/datasets/facebook/anli}{\nolinkurl{facebook/anli}} \\
HANS & Logic & \href{https://huggingface.co/datasets/jhu-cogsci/hans}{\nolinkurl{jhu-cogsci/hans}} \\
Pro-NoMath & Logic & \href{https://huggingface.co/datasets/sam-paech/mmlu-pro-nomath}{\nolinkurl{sam-paech/mmlu-pro-nomath}} \\
Temporal & Logic & \href{https://huggingface.co/datasets/Fhrozen/big_bench_hard}{\nolinkurl{Fhrozen/big_bench_hard}} \\
Tracking & Logic & \href{https://huggingface.co/datasets/Fhrozen/big_bench_hard}{\nolinkurl{Fhrozen/big_bench_hard}} \\
\midrule
GSM8K & Math & \href{https://huggingface.co/datasets/openai/gsm8k}{\nolinkurl{openai/gsm8k}} \\
MATH & Math & \href{https://huggingface.co/datasets/EleutherAI/hendrycks_math}{\nolinkurl{EleutherAI/hendrycks_math}} \\
OlympiadBench & Math & \href{https://huggingface.co/datasets/math-ai/olympiadbench}{\nolinkurl{math-ai/olympiadbench}} \\
Omni-MATH & Math & \href{https://huggingface.co/datasets/KbsdJames/Omni-MATH}{\nolinkurl{KbsdJames/Omni-MATH}} \\
\bottomrule
\end{tabularx}
\caption{Evaluation datasets and their Hugging Face repositories. ``Logic'' denotes Best-of-8 logical reasoning reranking, while ``Math'' denotes transfer evaluation on mathematical reasoning tasks.}
\label{tab:dataset-details}
\end{table}

\section{Training Details}

We fine-tune PRMs in an LM-head format with LoRA adaptation. The model receives a reasoning prefix and a candidate step, and assigns scores to the positive and negative label words. We use \texttt{true}/\texttt{false} as the default label words. The training objective is a pairwise ranking loss over the positive and negative label scores, so that valid steps are encouraged to receive higher scores than invalid ones.

For supervision construction, we use the \texttt{all\_after\_error} strategy: the first corrupted step and all following steps in the erroneous trajectory are treated as negative with respect to prefix-valid reasoning from the original problem state. Downstream steps are still recomputed under the corrupted state to preserve trajectory consistency, but they are labeled negative because their prefix has already become invalid. The best checkpoint is selected according to first-error accuracy on the validation set. All training runs are conducted on four NVIDIA RTX 4090 GPUs.

\begin{table}[t]
\centering
\small
\setlength{\tabcolsep}{3pt}
\begin{tabularx}{\columnwidth}{@{}l l X@{}}
\toprule
\textbf{Category} & \textbf{Parameter} & \textbf{Value} \\
\midrule
Model & Format & LM-head PRM \\
LoRA & Rank / alpha / dropout & 16 / 32 / 0.1 \\
\midrule
Objective & Loss & Pairwise ranking \\
Objective & Negative strategy & \texttt{all\_after\_error} \\
\midrule
Optimization & Optimizer & AdamW \\
Optimization & Learning rate & $2 \times 10^{-5}$ \\
Optimization & Weight decay & 0.01 \\
Optimization & Warmup ratio & 0.03 \\
Optimization & Epochs & 1 \\
\midrule
Batching & Max length & 1024 \\
Hardware & GPUs & $4 \times$ RTX 4090 \\
Selection & Metric & \texttt{first\_error\_acc} \\
\bottomrule
\end{tabularx}
\caption{Key training hyperparameters for PRM fine-tuning.}
\label{tab:training-details}
\end{table}

\section{Best-of-8 Generation, Evaluation Protocol, and Additional Results}
\label{app:best-of-8}

\begin{table*}[!t]
\centering
\small
\resizebox{\textwidth}{!}{%
\begin{tabular}{llcccccccc}
\toprule
\textbf{Family} & \textbf{Selector} & \textbf{folio} & \textbf{logicnli} & \textbf{pro\_nomath} & \textbf{anli} & \textbf{temporal} & \textbf{tracking} & \textbf{hans} & \textbf{avg} \\
\midrule
Llama & prm (ours) & 0.625 & 0.473 & 0.584 & 0.772 & 0.569 & 0.396 & 0.721 & 0.591 \\
Llama & Majority@8 & 0.525 & 0.387 & 0.470 & 0.790 & 0.720 & 0.560 & 0.811 & 0.574 \\
Llama & Oracle@8 & 0.900 & 0.852 & 0.722 & 0.933 & 0.604 & 0.696 & 0.948 & 0.860 \\
\midrule
Qwen & prm (ours) & 0.584 & 0.527 & 0.458 & 0.691 & 0.712 & 0.562 & 0.780 & 0.615 \\
Qwen & Majority@8 & 0.557 & 0.392 & 0.404 & 0.802 & 0.680 & 0.313 & 0.632 & 0.540 \\
Qwen & Oracle@8 & 0.702 & 0.802 & 0.583 & 0.909 & 0.824 & 0.812 & 0.968 & 0.800 \\
\bottomrule
\end{tabular}}
\caption{Dataset-level Best-of-8 comparison among our PRM, Majority@8, and Oracle@8. Candidate solutions are generated by the corresponding base model in each family.}
\label{tab:appendix-majority-oracle}
\end{table*}

\begin{table*}[!t]
\centering
\small
\resizebox{\textwidth}{!}{%
\begin{tabular}{lcccccccc}
\toprule
\textbf{Model} & \textbf{folio} & \textbf{logicnli} & \textbf{pro\_nomath} & \textbf{anli} & \textbf{temporal} & \textbf{tracking} & \textbf{hans} & \textbf{avg} \\
\midrule
llama-3.1-8B & 0.575 & 0.472 & 0.498 & 0.729 & 0.412 & 0.391 & 0.620 & 0.528 \\
llama-3.1-8B-5k & 0.582 & 0.436 & 0.558 & 0.758 & 0.461 & 0.374 & 0.652 & 0.545 \\
\midrule
\best{llama-3.1-8B-20k} & \best{0.625} & \best{0.473} & \best{0.584} & \best{0.772} & \best{0.569} & \best{0.396} & \best{0.721} & \best{0.591} \\
\bottomrule
\end{tabular}}
\caption{Best-of-8 reranking results on logical reasoning benchmarks under different training data scales. Candidate solutions are generated by the Llama-3.1-8B base model.}
\label{tab:appendix-scale-logic}
\end{table*}

For each test instance, we pair the input problem with eight sampled candidate solutions from the corresponding base model. During reranking, the PRM scores every candidate trajectory step by step. We aggregate step-level positive scores with the \texttt{min} rule, which makes the trajectory score sensitive to the weakest step in the reasoning chain. The highest-scoring trajectory is selected as the final prediction.

We also report two reference selectors. \textbf{Majority@8} selects the answer that appears most frequently among the eight sampled candidates. It provides a simple self-consistency baseline that does not use step-level supervision. \textbf{Oracle@8} checks whether the candidate pool contains at least one correct solution, and therefore estimates the upper bound imposed by the sampled candidates. The gap between our PRM and Oracle@8 reflects remaining headroom in step-level scoring and trajectory selection.

Table~\ref{tab:appendix-majority-oracle} reports the full dataset-level comparison with Majority@8 and Oracle@8. On both model families, our method improves over Majority@8 on average, suggesting that the learned process reward provides useful selection signals beyond answer-level voting. At the same time, the gap to Oracle@8 remains substantial, indicating considerable room for improvement in candidate selection.

\section{Effect of Training Data Scale}
\label{app:data-scale}

We further analyze the effect of synthesized data scale on logical reasoning. Table~\ref{tab:appendix-scale-logic} compares the base Llama-3.1-8B model with models trained using 5k and 20k synthesized instances. The average score rises from \textbf{0.528} for the base model to \textbf{0.545} with 5k training instances, and further to \textbf{0.591} with 20k instances. This trend suggests that the proposed synthesis framework continues to provide useful supervision as more process data are added. Although gains vary across datasets, the consistent average improvement indicates that scaling synthesized process data benefits PRM-based reranking.

\section{Error Type Distribution}
\label{app:error-distribution}
Table~\ref{tab:error-type-dist} summarizes the error-type distribution in the 20k training set. The \emph{Count} column reports the final number of instances used for training. We adjust the distribution to reduce the dominance of frequent template-induced errors and to make the training set more balanced across major error families. The \emph{Share} column is computed from these counts, which sum to 20,000.

The distribution is intentionally not perfectly uniform. We keep relatively frequent truth-state corruptions such as XOR-related errors and AND/OR confusion because they arise naturally from the symbolic rule templates, while retaining less frequent structural errors such as missing prerequisites and circular references to expose the model to trajectory-level failures.

\begin{table}[!t]
\centering
\small
\setlength{\tabcolsep}{4pt}
\begin{tabularx}{\columnwidth}{@{}Xrr@{}}
\toprule
\textbf{Error Type} & \textbf{Count} & \textbf{Share} \\
\midrule
\code{xor\_as\_equiv} & 3610 & 18.1\% \\
\code{xor\_as\_or} & 3609 & 18.0\% \\
\code{or\_and\_confusion} & 3598 & 18.0\% \\
\code{drop\_condition} & 1934 & 9.7\% \\
\code{implication\_misuse} & 1466 & 7.3\% \\
\code{converse\_error} & 1299 & 6.5\% \\
\code{redundant\_step} & 1185 & 5.9\% \\
\code{circular\_reference} & 946 & 4.7\% \\
\code{partial\_evaluation} & 913 & 4.6\% \\
\code{missing\_prerequisite} & 869 & 4.3\% \\
\code{vacuous\_truth\_error} & 571 & 2.9\% \\
\bottomrule
\end{tabularx}
\caption{Error-type distribution in the final 20k training set.}
\label{tab:error-type-dist}
\end{table}

\section{Error Type Definitions and Minimal Examples}
\label{app:error-definitions}

This section presents the error taxonomy used by our synthesis framework in a more structured form. We define errors from a \emph{truth-state satisfaction} perspective. A non-structural error either produces an assignment that violates the instantiated rule or current symbolic state, or performs a local update that is not licensed by the active rule. By contrast, a structural error may be locally truth-satisfying, but it violates the rule direction, the dependency order, or the well-formedness of the reasoning trajectory. This distinction matters for PRM supervision: a step may look locally plausible while still being invalid because the prefix does not support it.

\begin{table}[!t]
\centering
\small
\setlength{\tabcolsep}{3pt}
\begin{tabularx}{\columnwidth}{@{}lX@{}}
\toprule
\textbf{Group} & \textbf{Main criterion} \\
\midrule
Truth-state errors & The corrupted step either violates the local truth condition or current state, or performs an unauthorized local update. \\
Structural errors & The local conclusion may be truth-compatible, but the transition uses an invalid rule direction or an unsupported dependency. \\
\bottomrule
\end{tabularx}
\caption{High-level grouping of the error types used in our synthesized data.}
\label{tab:error-type-groups}
\end{table}

We use ASCII operators for readability: \code{and}, \code{or}, \code{xor}, and \code{->}. The examples below are minimal symbolic cases designed to clarify the boundary between categories.

\subsection{Local Truth-State Errors}

\begin{errorcard}{\code{drop\_condition}}
\errline{Definition}{A necessary condition in a compound premise is ignored. The corrupted step treats a partial match as if the full condition were satisfied.}
\validex{Given \code{[F2] -> ([F6] and [F7])}, \code{[F6]=False}, and \code{[F7]=True}, the correct update is \code{[F2]=False}.}
\badex{Given \code{[F2] -> ([F6] and [F7])}, \code{[F6]=False}, and \code{[F7]=True}, the corrupted step ignores \code{[F6]=False} and outputs \code{[F2]=True}.}
\end{errorcard}

\begin{errorcard}{\code{implication\_misuse}}
\errline{Definition}{The corrupted step violates the truth condition of implication. Under \code{A -> B}, the assignment \code{A=True, B=False} is invalid.}
\validex{Given \code{[F0] -> [F5]} and \code{[F5]=False}, the correct update is \code{[F0]=False}.}
\badex{Given \code{[F0] -> [F5]} and \code{[F5]=False}, the corrupted step outputs \code{[F0]=True}, which violates the implication constraint.}
\end{errorcard}

\begin{errorcard}{\code{or\_and\_confusion}}
\errline{Definition}{The step confuses the truth conditions of \code{and} and \code{or}.}
\validex{Given \code{[F7] -> ([F8] or [F5])}, \code{[F7]=True}, \code{[F8]=True}, and an existing state \code{[F5]=False}, the correct chain does not force \code{[F5]} to become true.}
\badex{Given the same rule and state, the corrupted chain treats the disjunction as if both branches must hold and incorrectly sets \code{[F5]=True}.}
\end{errorcard}

\begin{errorcard}{\code{partial\_evaluation}}
\errline{Definition}{Only part of a compound expression is evaluated, and this partial local observation is incorrectly treated as sufficient to determine another component.}
\validex{Given \code{[F6] -> ([F7] and [F3])}, \code{[F6]=True}, and \code{[F7]=True}, the correct update is \code{[F3]=True}.}
\badex{Given \code{[F6] -> ([F7] and [F3])}, \code{[F6]=True}, and \code{[F7]=True}, the corrupted step reasons only from one local component and outputs \code{[F3]=False}.}
\end{errorcard}

\begin{errorcard}{\code{xor\_as\_or}}
\errline{Definition}{XOR is treated as ordinary inclusive OR. The corrupted step incorrectly allows both sides to be true.}
\validex{Given \code{[F12] xor [F10]} and \code{[F12]=True}, the correct update is \code{[F10]=False}.}
\badex{Given \code{[F12] xor [F10]} and \code{[F12]=True}, the corrupted step outputs \code{[F10]=True}, incorrectly allowing both sides to hold.}
\end{errorcard}

\begin{errorcard}{\code{xor\_as\_equiv}}
\errline{Definition}{XOR is treated as equivalence. The corrupted step incorrectly forces both sides to have the same truth value.}
\validex{Given \code{[F3] xor [F0]} and \code{[F3]=False}, the correct update is \code{[F0]=True}.}
\badex{Given \code{[F3] xor [F0]} and \code{[F3]=False}, the corrupted step outputs \code{[F0]=False}, incorrectly making the two sides equivalent.}
\end{errorcard}

\begin{errorcard}{\code{vacuous\_truth\_error}}
\errline{Definition}{The step updates a consequent under a false antecedent. When the antecedent of \code{A -> B} is false, the implication is already satisfied and does not license changing the existing value of \code{B}.}
\validex{Given \code{[F3] -> [F4]}, \code{[F3]=False}, and an existing state \code{[F4]=True}, the correct update is to keep \code{[F4]=True}.}
\badex{Given \code{[F3] -> [F4]}, \code{[F3]=False}, and an existing state \code{[F4]=True}, the corrupted step overwrites the state and outputs \code{[F4]=False}.}
\end{errorcard}

\subsection{Trajectory-Structure Errors}

\begin{errorcard}{\code{converse\_error}}
\errline{Definition}{The step applies the converse direction of an implication even though the reversed rule is not present in the rule set.}
\validex{With the available rule \code{[F4] -> [F3]}, a valid use is to infer \code{[F3]=True} from \code{[F4]=True}.}
\badex{With only the rule \code{[F4] -> [F3]} available, the corrupted step uses the unavailable converse direction and derives \code{[F4]=True} from \code{[F3]=True}.}
\end{errorcard}

\begin{errorcard}{\code{redundant\_step}}
\errline{Definition}{A locally valid step is repeated after its conclusion has already been established in the prefix. Although the inference itself may be truth-compatible, it is invalid as a trajectory-structure error because it contributes no new supported progress.}
\validex{A useful step derives \code{[F3]=True} from \code{([F3] xor [F5]) -> [F6]}, \code{[F5]=True}, and \code{[F6]=False} when that conclusion is needed later in the chain.}
\badex{The corrupted step repeats the same derivation of \code{[F3]=True} after that fact has already been established and used.}
\end{errorcard}

\begin{errorcard}{\code{missing\_prerequisite}}
\errline{Definition}{A downstream rule is applied before a required bridge fact has been established in the prefix.}
\validex{First derive \code{[F5]=True} from \code{[F7] -> ([F8] or [F5])}; then use \code{([F3] xor [F5]) -> [F6]}, \code{[F5]=True}, and \code{[F6]=False} to derive \code{[F3]=True}.}
\badex{The corrupted step applies \code{([F3] xor [F5]) -> [F6]} directly and derives \code{[F3]=True} before \code{[F5]=True} has been established.}
\end{errorcard}

\begin{errorcard}{\code{circular\_reference}}
\errline{Definition}{Two or more steps mutually support each other instead of forming an acyclic forward chain.}
\validex{A valid forward derivation establishes each supporting fact before it is used in a later step, producing an acyclic dependency order.}
\badex{The corrupted pair forms a cycle: Step A derives \code{[F6]=True} using \code{[F1]=False}, while Step B derives \code{[F1]=False} using \code{[F6]=True}.}
\end{errorcard}

\section{Example Data Instances}
\label{app:examples}

The following examples illustrate how a synthesized instance stores both the correct and erroneous trajectories. We keep the step-level structure explicit because this is the supervision format used by the PRM: each step contains an applied rule, supporting facts, a conclusion, and, for the erroneous chain, the first invalid transition. The examples also show how the erroneous chain remains aligned with the correct chain while allowing a controlled deviation at the first-error step.

\subsection{Example A: Monkey Buns}

Monkey Buns grows up in a coastal town in a family bakery and becomes a cheerful local figure known for playful animal-shaped pastries. The target assignment is \code{[F0]=False}, rendered as ``Monkey Buns never took over the family bakery.'' The erroneous trajectory contains a \code{missing\_prerequisite} error: it applies a downstream rule before establishing the bridge fact \code{[F5]=True}.

\begin{examplecard}{Example A metadata}
\begin{tabularx}{\linewidth}{@{}lX@{}}
\textbf{Name} & Monkey Buns \\
\textbf{Goal} & ``Monkey Buns never took over the family bakery.'' \\
\textbf{Target assignment} & \code{[F0]=False} \\
\textbf{First error type} & \code{missing\_prerequisite} \\
\end{tabularx}
\end{examplecard}

\begin{examplecard}{Correct trajectory: Monkey Buns}
\begin{enumerate}[leftmargin=*,label=\textbf{Step \arabic*.},itemsep=0.45em,topsep=0.1em]
    \item \stepfield{Rule}{\code{[F9] xor [F12]}.}
    \stepfield{Facts}{\code{[F12]=False}; ``The idea of a dessert truck never crossed his mind.''}
    \stepfield{Conclusion}{\code{[F9]=True}; ``Monkey Buns personally delivers his pastries to customers.''}

    \item \stepfield{Rule}{\code{[F11] xor [F10]}.}
    \stepfield{Facts}{\code{[F11]=True}; ``Monkey Buns often shows up at children's parties with his pastries.''}
    \stepfield{Conclusion}{\code{[F10]=False}; ``He never dresses up as a monkey.''}

    \item \stepfield{Rule}{\code{([F9] and [F8]) -> [F10]}.}
    \stepfield{Facts}{\code{[F9]=True}; \code{[F10]=False}.}
    \stepfield{Conclusion}{\code{[F8]=False}; ``He's content to stay within the coastal town.''}

    \item \stepfield{Rule}{\code{[F7] -> ([F8] or [F5])}.}
    \stepfield{Facts}{\code{[F7]=True}; ``The townsfolk adore Monkey Buns as a local treasure''; \code{[F8]=False}.}
    \stepfield{Conclusion}{\code{[F5]=True}; ``He whips up pastries shaped like playful animals.''}

    \item \stepfield{Rule}{\code{([F3] xor [F5]) -> [F6]}.}
    \stepfield{Facts}{\code{[F5]=True}; \code{[F6]=False}; ``He often seems gloomy and downcast.''}
    \stepfield{Conclusion}{\code{[F3]=True}; ``Baking brings him pure joy.''}

    \item \stepfield{Rule}{\code{([F3] or [F4]) -> [F1]}.}
    \stepfield{Facts}{\code{[F3]=True}; \code{[F4]=False}; ``He struggles to keep people entertained.''}
    \stepfield{Conclusion}{\code{[F1]=True}; ``Monkey Buns spent his childhood in a small coastal town.''}

    \item \stepfield{Rule}{\code{([F0] and [F1]) -> [F2]}.}
    \stepfield{Facts}{\code{[F1]=True}; \code{[F2]=False}; ``No one in his family ever ran a bakery.''}
    \stepfield{Conclusion}{\code{[F0]=False}; ``Monkey Buns never took over the family bakery.''}
\end{enumerate}
\end{examplecard}

\begin{examplecard}{Erroneous trajectory: Monkey Buns}
\begin{enumerate}[leftmargin=*,label=\textbf{Step \arabic*.},itemsep=0.45em,topsep=0.1em]
    \item \stepfield{Rule}{\code{[F9] xor [F12]}.}
    \stepfield{Facts}{\code{[F12]=False}; ``The idea of a dessert truck never crossed his mind.''}
    \stepfield{Conclusion}{\code{[F9]=True}; ``Monkey Buns personally delivers his pastries to customers.''}

    \item \stepfield{Rule}{\code{[F11] xor [F10]}.}
    \stepfield{Facts}{\code{[F11]=True}; ``Monkey Buns often shows up at children's parties with his pastries.''}
    \stepfield{Conclusion}{\code{[F10]=False}; ``He never dresses up as a monkey.''}

    \item \stepfield{Rule}{\code{([F9] and [F8]) -> [F10]}.}
    \stepfield{Facts}{\code{[F9]=True}; \code{[F10]=False}.}
    \stepfield{Conclusion}{\code{[F8]=False}; ``He's content to stay within the coastal town.''}

    \item \stepfield{Rule}{\code{([F3] xor [F5]) -> [F6]}.}
    \stepfield{Facts}{\code{[F6]=False}; ``He often seems gloomy and downcast.''}
    \stepfield{Conclusion}{\code{[F3]=True}; ``Baking brings him pure joy.''}
    \firsterror{\code{missing\_prerequisite}. The required bridge fact \code{[F5]=True}, normally derived in the correct Step 4, has not been established.}

    \item \stepfield{Rule}{\code{([F3] or [F4]) -> [F1]}.}
    \stepfield{Facts}{\code{[F3]=True}; \code{[F4]=False}.}
    \stepfield{Conclusion}{\code{[F1]=True}; ``Monkey Buns spent his childhood in a small coastal town.''}

    \item \stepfield{Rule}{\code{([F0] and [F1]) -> [F2]}.}
    \stepfield{Facts}{\code{[F1]=True}; \code{[F2]=False}.}
    \stepfield{Conclusion}{\code{[F0]=False}; ``Monkey Buns never took over the family bakery.''}
\end{enumerate}
\end{examplecard}

\subsection{Example B: London}

London grows up in a foggy coastal town and later becomes a marine acoustician who studies whale songs and glacier sounds at sea. The target goal is ``London learned to navigate by sound long before they could see the path'' with symbolic value \code{[F1]=True}. The first error is \code{xor\_as\_equiv}, where the final XOR relation is incorrectly treated as equivalence.

\begin{examplecard}{Example B metadata}
\begin{tabularx}{\linewidth}{@{}lX@{}}
\textbf{Name} & London \\
\textbf{Goal} & ``London learned to navigate by sound long before they could see the path.'' \\
\textbf{Target value} & \code{[F1]=True} \\
\textbf{First error type} & \code{xor\_as\_equiv} \\
\end{tabularx}
\end{examplecard}

\begin{examplecard}{Correct trajectory: London}
\begin{enumerate}[leftmargin=*,label=\textbf{Step \arabic*.},itemsep=0.45em,topsep=0.1em]
    \item \stepfield{Rule}{\code{[F10] -> ([F11] and [F8])}.}
    \stepfield{Facts}{\code{[F10]=True}; ``London trusts their memory more than their eyes in the thick fog.''}
    \stepfield{Conclusion}{\code{[F8]=True}; ``London has the patience to wait for the ocean to break its long silences.''}

    \item \stepfield{Rule}{\code{[F4] xor [F9]}.}
    \stepfield{Facts}{\code{[F9]=False}; ``The soundscape of London's childhood has faded from memory.''}
    \stepfield{Conclusion}{\code{[F4]=True}; ``Part of London's work involves mapping the intricate songs of whales.''}

    \item \stepfield{Rule}{\code{[F6] xor [F8]}.}
    \stepfield{Facts}{\code{[F8]=True}.}
    \stepfield{Conclusion}{\code{[F6]=False}; ``London always waited for the sun to burn away the mist before venturing out.''}

    \item \stepfield{Rule}{\code{([F2] xor [F6]) -> [F7]}.}
    \stepfield{Facts}{\code{[F6]=False}; \code{[F7]=False}; ``The vibrations of the ocean remain imperceptible to London.''}
    \stepfield{Conclusion}{\code{[F2]=False}; ``London has a career that keeps them firmly on dry land.''}

    \item \stepfield{Rule}{\code{([F4] and [F3]) -> [F5]}.}
    \stepfield{Facts}{\code{[F4]=True}; \code{[F5]=False}; ``Fog was a rare and unfamiliar visitor in London's childhood.''}
    \stepfield{Conclusion}{\code{[F3]=False}; ``London's work rarely requires overnight trips away from shore.''}

    \item \stepfield{Rule}{\code{([F2] or [F0]) -> [F3]}.}
    \stepfield{Facts}{\code{[F2]=False}; \code{[F3]=False}.}
    \stepfield{Conclusion}{\code{[F0]=False}; ``London's childhood was spent far from any coastline.''}

    \item \stepfield{Rule}{\code{[F0] xor [F1]}.}
    \stepfield{Facts}{\code{[F0]=False}.}
    \stepfield{Conclusion}{\code{[F1]=True}; ``London learned to navigate by sound long before they could see the path.''}
\end{enumerate}
\end{examplecard}

\begin{examplecard}{Erroneous trajectory: London}
\begin{enumerate}[leftmargin=*,label=\textbf{Step \arabic*.},itemsep=0.45em,topsep=0.1em]
    \item \stepfield{Rule}{\code{[F10] -> ([F11] and [F8])}.}
    \stepfield{Facts}{\code{[F10]=True}; ``London trusts their memory more than their eyes in the thick fog.''}
    \stepfield{Conclusion}{\code{[F8]=True}; ``London has the patience to wait for the ocean to break its long silences.''}

    \item \stepfield{Rule}{\code{[F4] xor [F9]}.}
    \stepfield{Facts}{\code{[F9]=False}; ``The soundscape of London's childhood has faded from memory.''}
    \stepfield{Conclusion}{\code{[F4]=True}; ``Part of London's work involves mapping the intricate songs of whales.''}

    \item \stepfield{Rule}{\code{[F6] xor [F8]}.}
    \stepfield{Facts}{\code{[F8]=True}.}
    \stepfield{Conclusion}{\code{[F6]=False}; ``London always waited for the sun to burn away the mist before venturing out.''}

    \item \stepfield{Rule}{\code{([F2] xor [F6]) -> [F7]}.}
    \stepfield{Facts}{\code{[F6]=False}; \code{[F7]=False}; ``The vibrations of the ocean remain imperceptible to London.''}
    \stepfield{Conclusion}{\code{[F2]=False}; ``London has a career that keeps them firmly on dry land.''}

    \item \stepfield{Rule}{\code{([F4] and [F3]) -> [F5]}.}
    \stepfield{Facts}{\code{[F4]=True}; \code{[F5]=False}; ``Fog was a rare and unfamiliar visitor in London's childhood.''}
    \stepfield{Conclusion}{\code{[F3]=False}; ``London's work rarely requires overnight trips away from shore.''}

    \item \stepfield{Rule}{\code{([F2] or [F0]) -> [F3]}.}
    \stepfield{Facts}{\code{[F2]=False}; \code{[F3]=False}.}
    \stepfield{Conclusion}{\code{[F0]=False}; ``London's childhood was spent far from any coastline.''}

    \item \stepfield{Rule}{\code{[F0] xor [F1]}.}
    \stepfield{Facts}{\code{[F0]=False}.}
    \stepfield{Conclusion}{\code{[F1]=False}; ``London did not learn to navigate by sound before they could see the path.''}
    \firsterror{\code{xor\_as\_equiv}. This step treats XOR as equivalence, assigning the other branch the same truth value as \code{[F0]} rather than the correct opposite value.}
\end{enumerate}
\end{examplecard}

\section{Prompt Templates}
\label{app:prompts}

This section summarizes the prompts used in PRM training, Best-of-8 reranking, and natural-language translation. We present cleaned versions of the implementation prompts in a compact dialogue format. Placeholders such as \code{\{goal\}}, \code{\{question\}}, and \code{\{current\_step\}} are filled automatically by the pipeline. Control fields such as step status and error type are used only during offline data synthesis and annotation; they are not included as label-leaking text in the PRM input. The prompts are grouped by function rather than by source file for readability.

\subsection{Step-Level PRM Prompts}

\paragraph{Training prompt.}
During fine-tuning, each symbolic trajectory is expanded into step-level instances. The model receives the target goal, initial information, previous steps, and the current step, and is trained to generate one of two label words. In the main experiments, the label words are \code{true} and \code{false}.

\begin{promptbox}{PRM training prompt}
\promptrole{User}
Task: Determine whether the current reasoning step is logically correct.

\textbf{Goal:} \code{\{goal\}}

\textbf{Initial Information:} \code{\{initial\_information\}}

\textbf{Previous Steps:} \code{\{previous\_steps\}}

\textbf{Current Step:} \code{\{current\_step\}}

\textbf{Answer:}

\promptsep
\promptrole{Assistant}
\code{true} or \code{false}
\end{promptbox}

Each step is serialized with its supporting facts, the applied rule, and the resulting conclusion. An auxiliary reasoning field can also be appended, though it is not required in the main experiments. The same dialogue format is used for both correct and erroneous chains; the label is determined by whether the current step is valid under its prefix.

\paragraph{Best-of-8 scoring prompt.}
At evaluation time, each sampled candidate solution is split into steps. For the $t$-th step, the prompt contains the original question, the candidate prefix before that step, and the current step. The PRM scores the probability of the positive label word, and the final trajectory score is obtained by aggregating step-level scores with the \code{min} rule.

\begin{promptbox}{Best-of-8 step scoring prompt}
\promptrole{User}
Task: Determine whether the current reasoning step is logically correct.

\textbf{Question:} \code{\{question\}}

\textbf{Previous Steps:} \code{\{previous\_steps\}}

\textbf{Current Step:} \code{\{current\_step\}}

\textbf{Answer:}

\promptsep
\promptrole{Assistant}
\code{true} or \code{false}
\end{promptbox}

This prompt differs from the training prompt mainly in its context field: it uses the original problem question instead of the translated symbolic goal. The previous steps are the candidate's own generated prefix, so the judge evaluates whether the current step remains valid in that candidate-specific context.

\subsection{Natural-Language Translation Prompts}

The translation pipeline converts symbolic PRM instances into aligned natural-language data. All stages share the same entity name, background story, and predicate mapping for the correct and erroneous chains. This shared semantic space keeps paired trajectories comparable: their differences should reflect reasoning discrepancies rather than unrelated wording changes.

\paragraph{Background generation.}
The first stage creates a short scenario for the sampled entity. This scenario provides a coherent semantic context for all predicates in the same instance.

\begin{promptbox}{Background generation}
\promptrole{System}
Generate a brief background story for a person. The story should be natural, concise, and provide context for logical reasoning about the person's attributes and actions. Output only the story.

\promptsep
\promptrole{User}
Generate a background story for someone named \code{\{name\}}.

\promptsep
\promptrole{Assistant}
\code{\{background\_story\}}
\end{promptbox}

\paragraph{Predicate mapping.}
The second stage maps fact symbols to natural-language predicates. For each symbol, the model must produce a predicate name, a positive realization, and a negative realization. This allows the translator to render truth values consistently across the goal, base facts, rules, and both chains.

\begin{promptbox}{Predicate mapping prompt}
\promptrole{System}
Convert logical fact symbols such as \code{[F0]} and \code{[F1]} into vivid, natural English predicates that fit the character's background.

Return valid JSON with one entry per fact symbol. Each entry should include a predicate name, a natural sentence for the true value, and a natural sentence for the false value.

\smallskip
\textbf{Requirements:}
each predicate must be unique; sentence structures should be varied; negative forms should sound natural rather than mechanical; the background story should guide the predicates; output only valid JSON, with no markdown or extra text.

\promptsep
\promptrole{User}
\textbf{Background:} \code{\{background\}}

\textbf{Name:} \code{\{name\}}

\textbf{Fact symbols:} \code{\{fact\_symbols\}}

Create unique, natural-sounding predicates for each fact symbol. Output only JSON.

\promptsep
\promptrole{Assistant}
\code{\{predicate\_mapping\_json\}}
\end{promptbox}

\paragraph{Expression and rule translation.}
Simple facts are translated by selecting the corresponding positive or negative realization from the predicate mapping. Compound expressions and symbolic rules are translated with separate prompts conditioned on the predicate mapping. The rule prompt encourages varied realizations of implication, conjunction, disjunction, and exclusive-or relations, rather than repeatedly using rigid ``if ... then ...'' templates.

\begin{promptbox}{Expression and rule translation prompts}
\promptrole{System: expression translation}
Translate the logical expression and its truth value into natural English. Use the provided predicate mapping. Output only the natural-language sentence.

\promptrole{User: expression translation}
\textbf{Expression:} \code{\{expression\}}

\textbf{Truth value:} \code{\{truth\_value\}}

\textbf{Predicate mapping:} \code{\{predicate\_mapping\}}

\textbf{Name:} \code{\{name\}}

Translate the expression into natural English.

\promptsep
\promptrole{System: rule translation}
Translate the logical rule into natural, conversational English. Vary the sentence structure.

For implication, use patterns such as ``When X, Y follows'' or ``X means Y.'' For disjunction, express that at least one option holds. For conjunction, express that both statements hold. For exclusive-or, express that exactly one side is true and the two sides cannot both be true.

Treat \code{[[F]]} the same as \code{[F]}. Output only the translated rule.

\promptrole{User: rule translation}
\textbf{Rule:} \code{\{rule\}}

\textbf{Predicate mapping:} \code{\{predicate\_mapping\}}

\textbf{Name:} \code{\{name\}}

Translate the rule into natural, conversational English.

\promptsep
\promptrole{Assistant}
\code{\{expression\_nl\}} or \code{\{rule\_nl\}}
\end{promptbox}

\paragraph{Step-level translation and refinement.}
The final stage translates each symbolic step into a natural-language reasoning step. For correct steps, the text states why the conclusion follows from the facts and rule. For erroneous steps, the translator may produce an explanatory annotation of the process problem, but the text used as PRM input is further naturalized to avoid revealing the label. Structural errors are handled by focusing on missing prerequisites, invalid rule direction, redundant step reuse, or circular dependencies rather than simply treating the conclusion as false.

This prompt is used only during offline natural-language synthesis and annotation. Control fields such as step status and error type guide the translator, but they are not included in the final PRM input. For error-chain steps from the first invalid step onward, the naturalized text avoids explicit markers such as ``error'', ``mistake'', ``wrong'', ``invalid'', and similar meta-commentary. This preserves the intended propositions while keeping the generated process usable for PRM training.

\begin{promptbox}{Step-level translation and refinement prompt}
\promptrole{System}
Translate the reasoning step into concise, natural English.

Keep the statement faithful to the given facts, rule, conclusion, and step status. Output only the step text.

\promptsep
\promptrole{User}
\textbf{Background:} \code{\{background\}}

\textbf{Name:} \code{\{name\}}

\textbf{Facts:} \code{\{facts\_nl\}}

\textbf{Rule:} \code{\{rule\_nl\}}

\textbf{Conclusion:} \code{\{conclusion\_nl\}}

\textbf{Step status:} \code{\{correct\_or\_erroneous\}}

\textbf{Error type:} \code{\{error\_type\}}

Write one or two natural sentences for this reasoning step. Preserve all truth values and propositions.

If the step is correct, briefly state why the conclusion fits the facts and rule.

If the output is used as an annotation for an erroneous step, describe the process problem clearly.

If the output is used as PRM input, naturalize the wording and avoid explicit label-leaking words such as ``error'', ``mistake'', ``wrong'', ``invalid'', ``unsupported'', ``evidence'', ``established'', ``assumes'', ``depends'', ``relies'', ``repeats'', or ``restates''.

For structural errors, focus on the missing prerequisite, rule-direction, redundant-step, or cyclic-dependency issue rather than merely saying that the conclusion is false.

Avoid meta phrases such as ``the rule says'', ``according to the rule'', ``this step'', or ``the conclusion''.

Keep the wording concise and conversational.

\promptsep
\promptrole{Assistant}
\code{\{step\_reasoning\_nl\}}
\end{promptbox}

The same step-level translation procedure is applied to both correct and erroneous chains under the shared semantic mapping. The resulting paired trajectories remain aligned at the level of facts, rules, and conclusions, while their differences reflect the intended reasoning discrepancy.

\end{document}